\documentclass[conference]{IEEEtran}

\usepackage{cite}
\usepackage{amsmath,amssymb,amsfonts}
\usepackage{booktabs}
\usepackage{multirow}
\usepackage{algorithmic}
\usepackage{graphicx}
\usepackage{textcomp}
\usepackage{xcolor}
\usepackage{hyperref}

\def\BibTeX{{\rm B\kern-.05em{\sc i\kern-.025em b}\kern-.08em
    T\kern-.1667em\lower.7ex\hbox{E}\kern-.125emX}}
\begin{document}

\makeatletter
\def\ps@IEEEtitlepagestyle{%
  \def\@oddhead{\normalfont\footnotesize\sffamily\hfil Accepted to the Workshop on Data Quality Aware, High-Performance, and Trustworthy AI Systems for Healthcare at IEEE/ACM CHASE 2026\hfil}%
  \def\@evenhead{}%
  \def\@oddfoot{\normalfont\footnotesize\sffamily
    \begin{minipage}{\textwidth}
    \centering
    © Owner/Author 2026. This is the author's version of the work. It is posted here for your personal use. Not for redistribution. \\ 
    The definitive Version of Record will be published in \textit{Proceedings of the IEEE/ACM CHASE 2026}
    \end{minipage}}%
  \def\@evenfoot{}%
}
\makeatother

\title{Spatially Grounded Concept Bottleneck Models for Trustworthy Breast Ultrasound Diagnosis}

\author{
\IEEEauthorblockN{
Moshiur Rahman Tonmoy\IEEEauthorrefmark{1},
Dunren Che\IEEEauthorrefmark{2},
Haitham Y. Adarbah\IEEEauthorrefmark{3},
and Afzel Noore\IEEEauthorrefmark{4}
}
\IEEEauthorblockA{
Department of Electrical Engineering and Computer Science, Texas A\&M University--Kingsville (TAMUK),\\
Kingsville, TX 78363, USA
}
\IEEEauthorblockA{
\IEEEauthorrefmark{1}\texttt{Moshiur\_Rahman.Tonmoy@students.tamuk.edu},
\IEEEauthorrefmark{2}\texttt{Dunren.Che@tamuk.edu}\\
\IEEEauthorrefmark{3}\texttt{Haitham.Adarbah@tamuk.edu},
\IEEEauthorrefmark{4}\texttt{Afzel.Noore@tamuk.edu}
}
}

\maketitle

\begin{abstract}
Concept Bottleneck Models provide interpretable-by-design predictions by mediating diagnosis through human-understandable concepts, but in medical imaging, their trustworthiness is often limited by the quality and granularity of available supervision. In particular, predicted concept activations can be driven by irrelevant regions, leading to spatially unfaithful explanations. We study a data-centric spatially grounded Concept Bottleneck Model (SG-CBM) that leverages coarse lesion delineations as weak supervision to encourage anatomically plausible concept evidence. For breast ultrasound, we derive two clinically motivated zones from each lesion mask: (i) an in-lesion region of interest for morphology-related concepts and (ii) a posterior acoustic band for posterior phenomena. We train concept maps using a grouped spatial grounding objective and preserve semantic faithfulness with a linear bottleneck classifier. Across five-fold stratified group cross-validation, the proposed SG-CBM improves diagnostic AUROC and concept macro-AUROC while markedly increasing spatial alignment of concept evidence. We also perform a Train-corrupt/Test-clean annotation-quality stress test to quantify the impact of supervision quality on diagnosis and spatial faithfulness. Overall, the results underscore the need for data-quality-aware supervision design and systematic trustworthiness validation for deployable healthcare AI systems.

\end{abstract}


\section{Introduction}\label{sec:intro}
Recent advances in deep learning (DL) and computer vision have significantly improved medical imaging analysis, enabling automated detection, segmentation, and classification of breast lesions \cite{WANG2025100138}. Despite these advances, the black-box nature of many DL models raises concerns about reliability, transparency, and clinical trustworthiness \cite{Kondylakis2025Review}. In safety-critical settings such as healthcare, clinicians must understand why a model reaches a particular decision. Two related notions address this need: \emph{explainability} and \emph{interpretability}. Although often used interchangeably, they describe different forms of model transparency \cite{RETZLAFF2024101243}. Explainability typically refers to post-hoc techniques that interpret model outputs after training, such as saliency maps or heatmaps \cite{TONMOY2025xbrain}. While such explainable AI (XAI) approaches provide insights into model behavior, they do not guarantee that predictions rely on clinically relevant evidence. Models may still exhibit the so-called Clever Hans behavior, achieving high accuracy while relying on spurious correlations, artifacts, or background cues rather than true lesion characteristics \cite{Mustafa2024unmasking, vasquez2025detecting}. This undermines generalizability and limits clinical adoption.

Interpretability, in contrast, refers to ante-hoc transparency, where the model structure and decision process are inherently understandable. Concept Bottleneck Models (CBMs) \cite{koh2020CBM} provide such an interpretable-by-design framework by mediating predictions through human-understandable concepts. In a CBM, the model first predicts clinical concepts and then derives the final diagnosis solely from these predictions, enabling structured explanations. For example, Bunnell et al. \cite{Bun_Learning_MICCAI2024} integrate BI-RADS (Breast Imaging-Reporting and Data System) descriptors within a concept bottleneck layered on a Mask R-CNN backbone to produce interpretable malignancy predictions from breast ultrasound. Other studies also incorporate BI-RADS descriptors in DL pipelines; for instance, Zhang et al. \cite{Zhang2021BiRadsNet} proposed BI-RADS-NET, a multitask learning framework that predicts descriptors alongside diagnosis, while Carrilero-Mardones et al. \cite{Mardones2024Deep} introduced a joint detection-description-classification pipeline using breast ultrasound and BI-RADS descriptors. Although these approaches enhance interpretability, the absence of a bottleneck layer means predictions are not strictly mediated by BI-RADS descriptors.

Beyond model choice, healthcare AI trustworthiness is constrained by the quality, granularity, and consistency of its data and supervision signals. In breast ultrasound, concept annotations (e.g., BI-RADS descriptors) are typically provided at the lesion level, and the lesion delineation may be available or obtainable \cite{review2025Mudassar}, while pixel-level annotations that localize each concept are rarely available. This supervision-granularity mismatch becomes particularly salient for CBMs: although diagnosis is mediated by predicted concepts, the concept predictors may still lack spatial grounding, predicting the correct concepts while allowing activations to attend to irrelevant regions, artifacts, or dataset biases and yielding spatially unfaithful explanations, limiting trust and auditability \cite{Huang_Song_Hu_Zhang_Wang_Song_2024AAAI,knab2026whats}. In principle, pixel-level concept annotations could enforce spatial grounding, but such fine-grained supervision is costly and seldom feasible in practice.

Prior work shows that weakly supervised learning can leverage coarse annotations in breast ultrasound while improving lesion localization and diagnostic reliability \cite{Wen2025Identification, Wang2024Weakly}. Motivated by this, we propose \textbf{SG-CBM}, a spatially grounded CBM for trustworthy breast ultrasound diagnosis. During training, predicted concept maps are guided by a grouped weak spatial supervision objective that encourages each concept to activate within clinically appropriate regions while suppressing responses elsewhere. We define two clinically motivated spatial zones: (i) an in-lesion region-of-interest (ROI) corresponding to the lesion area, suitable for morphology-related concepts such as shape and margin, and (ii) a posterior acoustic band beneath the lesion that captures posterior enhancement or shadowing phenomena. These zones reflect radiological reasoning while avoiding pixel-level concept annotations. Importantly, we view this zone-based supervision as a form of task-driven data quality assurance: rather than only optimizing predictive accuracy, we enforce that the evidence supporting each concept arises from anatomically plausible regions. This enables more reliable trustworthiness validation because concept explanations can be audited both semantically, by assessing which concepts are present, and spatially, by assessing where the supporting evidence is located.

As a result, the learned concept representations remain structurally faithful and spatially aligned with clinically relevant regions. The main contributions of this work can be summarized as:
\begin{itemize}
    \item We introduce SG-CBM, a zone-aware spatial grounding framework for CBMs in ultrasound.
    \item We propose a lesion-based weak supervision strategy that enables spatial grounding without pixel-level concept annotations.
    \item We perform a structured evaluation across diagnosis, concept prediction, and spatial grounding.

    \item We further study supervision data quality via a Train-corrupt/Test-clean annotation-quality stress test, quantifying how training-time mask degradation affects diagnostic performance, concept quality, and spatial faithfulness.
\end{itemize}

\section{Methodology}\label{sec:methodology}

\begin{figure}
\centering
\includegraphics[width=\linewidth, trim=1.0cm 0.50cm 0.60cm 0.60cm, clip]{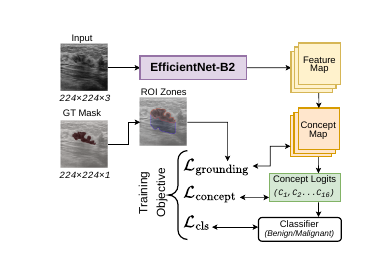}
\caption{High-level architecture of SG-CBM. The ROI zone (red) and posterior band (blue) are derived from the ground-truth lesion mask and used to provide weak spatial supervision through the grounding loss.}
\label{fig:framework}
\end{figure}

We employ lesion delineation as a coarse supervision signal to derive clinically motivated zones and impose a task-driven data quality assurance constraint that aligns concept evidence with these zones during training. Specifically, we treat off-zone activation as an evidence-quality violation and penalize it through separation and mass concentration losses. This yields a data-centric mechanism to improve the reliability of concept-based explanations without requiring concept-specific dense spatial labels. Let $\mathcal{D}=\{(x_i,y_i,c_i,m_i)\}_{i=1}^N$ denote a breast ultrasound dataset containing grayscale images $x\in\mathbb{R}^{224\times224\times1}$, diagnosis labels $y\in\{0,1\}$, $K=16$ binary clinical concepts $c\in\{0,1\}^{K}$, and lesion masks $m\in\{0,1\}^{224\times224}$.  
Images are replicated to three channels and processed by an ImageNet-pretrained EfficientNet-B2 backbone. An intermediate feature tensor $F\in\mathbb{R}^{C\times H_c\times W_c}$ ($14\times14\times112$) is mapped to $K=16$ concept maps $S=g_\phi(F)$ via a $1\times1$ convolution.  
Scalar concept logits $\hat{s}_k$ are obtained using top-$25\%$ {average} pooling over each map, producing {sigmoid} concept probabilities $\hat{p}=\sigma(\hat{s})$.  
Diagnosis is predicted solely from the concept vector using a linear classifier $\hat{y}=h_\psi(\hat{p})$. Figure \ref{fig:framework} illustrates the high-level workflow of the SG-CBM framework.

\subsection{Spatial grounding zones}
Spatial supervision is derived from the lesion mask $m$, and the two anatomical zones are defined as follows:

\paragraph{ROI Zone ($Z_{ROI}$)}
The lesion region provides a natural anatomical prior for morphology-related concepts. To reduce sensitivity to minor boundary errors and to compensate for resolution mismatch between image-space masks and lower-resolution concept maps, the $Z_{ROI}$ is obtained by dilating the $m$ using \emph{max-pooling} with kernel size $5$ for $1$ iteration.

\paragraph{Posterior Band ($Z_{POST}$)}

Posterior acoustic phenomena typically manifest directly beneath a lesion; however, supervising the entire region below the lesion can introduce noisy supervision, since deeper tissue often contains irrelevant structures and imaging patterns. Moreover, posterior acoustic features are not consistently present or visually salient in all malignant ultrasound scans, but for simplicity, we assume that a posterior band may be informative across cases in this work.
We therefore define a size-adaptive posterior band immediately beneath the lesion to provide a more stable spatial prior for posterior concepts. For each image column $u$, let $v_{\max}(u)$ denote the lowest lesion pixel. The posterior band extends vertically below this boundary:
\begin{equation}
Z_{\mathrm{POST}}(v,u)=
\begin{cases}
1 & v \in [\,v_{\max}(u)+1,\; v_{\max}(u)+H_b\,] \\
0 & \text{otherwise}.
\end{cases}
\end{equation}
The band height is defined adaptively from lesion size: $H_b=\min\!\left(H_{\max},\;\max\!\left(H_{\min},\;\lceil \alpha\,h_{\mathrm{bbox}}\rceil\right)\right)$, where $h_{\mathrm{bbox}}$ is the lesion bounding-box height. We set $\alpha=1.0$, $H_{\min}=10$, and $H_{\max}=50$ pixels.

Next, both zones are resized to concept-map resolution:
$\tilde{Z}_{\mathrm{ROI}},\tilde{Z}_{\mathrm{POST}}\in\{0,1\}^{H_c\times W_c}$. ROI-related concepts (shape, margin, echo pattern) are grouped to $\tilde{Z}_{\mathrm{ROI}}$, while posterior acoustic concepts are assigned to $\tilde{Z}_{\mathrm{POST}}$.

\subsection{Training objective}
Concept prediction is supervised using weighted BCE, and diagnosis is supervised with focal loss \cite{Lin2017Focal}. To enforce spatial faithfulness as an evidence quality constraint, concept activations $A=\sigma(S)$ are encouraged to concentrate within their assigned zones. For a zone mask $\tilde{Z}$, define the average activation inside and outside the zone:
\begin{equation}
\mu_{\mathrm{in}}(k)=\frac{\sum A_k \tilde{Z}}{\sum \tilde{Z}+\epsilon};
\quad
\mu_{\mathrm{out}}(k)=\frac{\sum A_k(1-\tilde{Z})}{\sum (1-\tilde{Z})+\epsilon}.
\end{equation}
We use a separation loss encouraging higher activation inside the target zone:
\begin{equation}
\mathcal{L}_{\mathrm{sep}}=\frac{1}{|\mathcal{K}|}\sum_{k\in\mathcal{K}}
\max\bigl(0,\;\delta-(\mu_{\mathrm{in}}(k)-\mu_{\mathrm{out}}(k))\bigr),
\end{equation}
and a mass concentration loss, encouraging most activation mass to lie inside the zone:
\begin{equation}
\mathcal{L}_{\mathrm{mass}}=\frac{1}{|\mathcal{K}|}\sum_{k\in\mathcal{K}}
\left(1-\frac{\sum A_k\tilde{Z}}{\sum A_k+\epsilon}\right).
\end{equation}
The zone loss is:
\begin{equation}
\mathcal{L}_{\mathrm{zone}}=\lambda_{\mathrm{sep}}\mathcal{L}_{\mathrm{sep}}+
\lambda_{\mathrm{mass}}\mathcal{L}_{\mathrm{mass}},
\end{equation}
with $\lambda_{\mathrm{sep}}=\lambda_{\mathrm{mass}}=1$, $\epsilon = 10^{-6}$, and margin $\delta=10\%$. Next, instead of uniform averaging across zones, we weigh each zone by its concept count so that each concept contributes comparably. Let $n_{\mathrm{ROI}}$ and $n_{\mathrm{POST}}$ denote the number of concepts in each group:
\begin{equation}
\mathcal{L}_{\mathrm{grounding}}
=
\frac{n_{\mathrm{ROI}}}{n_{\mathrm{ROI}}+n_{\mathrm{POST}}}\mathcal{L}^{\mathrm{ROI}}_{\mathrm{zone}}
+
\frac{n_{\mathrm{POST}}}{n_{\mathrm{ROI}}+n_{\mathrm{POST}}}\mathcal{L}^{\mathrm{POST}}_{\mathrm{zone}}
\end{equation}

The final training objective is:
\begin{equation}
\mathcal{L}
=
\lambda_c\mathcal{L}_{\mathrm{concept}}
+
\lambda_y\mathcal{L}_{\mathrm{cls}}
+
\lambda_s\mathcal{L}_{\mathrm{grounding}},
\end{equation}
with $\lambda_c=1$, $\lambda_y=1$, and $\lambda_s=2$.

\section{Experiments and Results}\label{sec:results}

\subsection{Testbed and Evaluation}
We treat breast ultrasound diagnosis on the \emph{BrEaST} dataset \cite{Pawlowska2024, Pawlowska2024BreastLesionsUSG} as a case study demonstrating how supervision granularity and mask-derived priors influence trustworthy concept evidence. The dataset provides grayscale breast ultrasound images with lesion masks and BI-RADS descriptors. Images and masks are resized to $224\times224$. From the BI-RADS annotations, we curate $K=16$ binary clinical concepts grouped into $\tilde{Z}_{\mathrm{ROI}}$ (shape, margin, echogenicity descriptors) and $\tilde{Z}_{\mathrm{POST}}$ (enhancement, shadowing). Evaluation uses 5-fold stratified group cross-validation. The best checkpoint is selected by validation AUROC, and results are reported as mean $\pm$ standard deviation (std) across folds. Training uses AdamW with cosine learning-rate scheduling. We first train the concept head and classifier for 10 epochs with the backbone being frozen, followed by 100 epochs of end-to-end training with spatial grounding enabled. Data augmentation includes horizontal flipping, brightness/contrast adjustment, Gaussian noise, and random rotations ($\pm15^\circ$). 

We evaluate SG-CBM along three complementary axes: (i) \emph{clinical utility} (diagnostic fidelity), (ii) \emph{semantic interpretability} (concept quality), and (iii) \emph{spatial faithfulness} ({whether concept evidence lies within predefined clinically meaningful zones}). For quantifying spatial grounding, we utilize and present three zone-based metrics:
\begin{itemize}
    \item \textbf{Energy-in-Zone:}
        \begin{equation}
        \mathrm{Energy}(k)=
        \frac{\sum_{i=1}^{N} A_k(i)\,\tilde{Z}(i)}
        {\sum_{i=1}^{N} A_k(i) + \epsilon}
        \end{equation}
    {It measures the fraction of activation mass within the zone; therefore, higher values indicate that the model's spatial evidence for concept $k$ is concentrated in the expected anatomical region, supporting clinically plausible reasoning.}
    \item \textbf{Hit@1:}
        \begin{equation}
        \mathrm{Hit@1}(k)=
        \tilde{Z}\!\left(\arg\max_{i\in\{1,\dots,N\}} A_k(i)\right)
        \end{equation}
    {This verifies whether the model's strongest spatial evidence for the concept is located in the clinically relevant zone, which is important for assessing whether the most decisive evidence is spatially meaningful.}
    \item \textbf{Top-$5\%$ Overlap:} {extends Hit@1 by measuring the fraction of the top $5\%$ activated pixels located within the zone, instead of checking only the single most activated pixel. High overlap suggests that the model's strongest evidence is consistently localized to the expected anatomical region rather than isolated or spurious pixels.}
\end{itemize}
Metrics are computed per concept and averaged across concept groups and folds for reporting.


\subsection{Results}\label{subsec:results}

\begin{table*}[htbp] 
\caption{ Comparison of predictive performance, concept quality, and spatial grounding. \emph{Vanilla CBM} denotes a CBM without spatial grounding. Concept quality is measured using macro AUROC over 16 concepts. Grounding quality is evaluated using Energy-in-Zone, Hit@1, and TopK Overlap for $\tilde{Z}_{\mathrm{ROI}}$ and $\tilde{Z}_{\mathrm{POST}}$. Values are mean $\pm$ std over 5 folds. }
\centering 
\small 
\setlength{\tabcolsep}{4pt} 
\renewcommand{\arraystretch}{1.15} 
\resizebox{\textwidth}{!}{ \begin{tabular}{l cc c cccccc} 
\toprule 
\multirow{3}{*}{Model} & \multicolumn{2}{c}{Diagnosis} & Concepts & \multicolumn{6}{c}{Grounding} \\ 
\cmidrule(lr){2-3} \cmidrule(lr){5-10} 
& \multirow{2}{*}{Accuracy} & \multirow{2}{*}{AUROC} & \multirow{2}{*}{AUROC} & \multicolumn{2}{c}{Energy} & \multicolumn{2}{c}{Hit@1} & \multicolumn{2}{c}{Top-5\% Overlap} \\ 
\cmidrule(lr){5-6} \cmidrule(lr){7-8} \cmidrule(lr){9-10} 
&&& & ROI & Post & ROI & Post & ROI & Post \\ 
\midrule 
EfficientNet-B2 & $0.751\pm0.064$ & $0.874\pm0.023$ & -- & -- & -- & -- & -- & -- & -- \\ 
Vanilla CBM & $0.746\pm0.030$ & $0.869\pm0.040$ & $0.741\pm0.037$ & $0.149\pm0.023$ & $0.099\pm0.020$ & $0.211\pm0.043$ & $0.091\pm0.011$ & $0.171\pm0.029$ & $0.096\pm0.017$ \\ 
\textbf{SG-CBM} & $\mathbf{0.802\pm0.034}$ & $\mathbf{0.892\pm0.012}$ & $\mathbf{0.771\pm0.023}$ & $\mathbf{0.815\pm0.071}$ & $\mathbf{0.585\pm0.165}$ & $\mathbf{0.915\pm0.046}$ & $\mathbf{0.661\pm0.178}$ & $\mathbf{0.689\pm0.052}$ & $\mathbf{0.567\pm0.103}$ \\ 
\bottomrule \end{tabular} 
} 
\label{tab:main_results} 
\end{table*}

\begin{figure*}
    \centering
    \includegraphics[width=\linewidth, trim=0.00cm 17.90cm 0.00cm 0.00cm, clip]{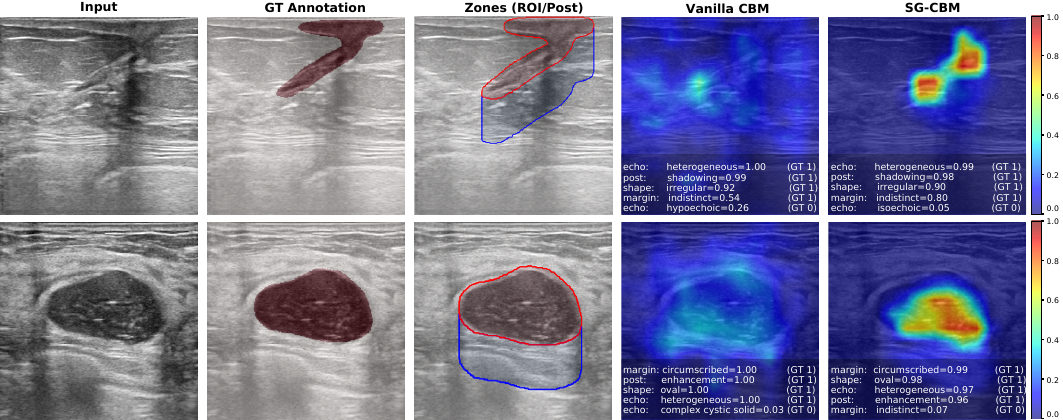}
    \caption{Qualitative comparison of spatial grounding, illustrating the averaged sigmoid activations of ROI and posterior-related concepts on the rightmost 2 panels. SG-CBM exhibits stronger spatial concentration within anatomically consistent regions, consistent with the quantitative grounding metrics.}
    \label{fig:localization_comparision}
\end{figure*}

\begin{table*}[htbp] 
\caption{Performance under training-time mask erosion (Train-corrupt/Test-clean). Models are trained with eroded lesion masks (kernel $K$) used to derive grounding zones; all evaluation metrics (including grounding) are computed using zones derived from the clean masks. Values are mean±std over 5 folds.}
\centering 
\small 
\setlength{\tabcolsep}{4pt} 
\renewcommand{\arraystretch}{1.15} 
\resizebox{\textwidth}{!}{ \begin{tabular}{ccc c cccccc} 
\toprule 
Erosion & Quality & Diagnosis & Concepts & \multicolumn{6}{c}{Grounding} \\ \cmidrule(lr){5-10} 
\multirow{2}{*}{$K$}& \multirow{2}{*}{Dice} & \multirow{2}{*}{AUROC} & \multirow{2}{*}{AUROC} & \multicolumn{2}{c}{Energy} & \multicolumn{2}{c}{Hit@1} & \multicolumn{2}{c}{Top-5\% Overlap} \\ 
\cmidrule(lr){5-6} \cmidrule(lr){7-8} \cmidrule(lr){9-10} 
&&& & ROI & Post & ROI & Post & ROI & Post \\ 
\midrule 
- & $1.000\pm0.000$ & $\mathbf{0.892\pm0.012}$ & $\mathbf{0.771\pm0.023}$ & $0.815\pm0.071$ & $0.585\pm0.165$ & $0.915\pm0.046$ & $0.661\pm0.178$ & $\mathbf{0.689\pm0.052}$ & $0.567\pm0.103$ \\ 

3 & $0.937\pm0.002$ & $0.853\pm0.028$ & $\underline{0.768\pm0.006}$ & $0.876\pm0.088$ & $\mathbf{0.690\pm0.139}$ & $\underline{0.935\pm0.055}$ & $\mathbf{0.782\pm0.122}$ & \underline{$0.646\pm0.052$} & $\mathbf{0.624\pm0.077}$ \\

5 & $0.870\pm0.004$ & $0.878\pm0.028$ & $0.764\pm0.015$ & \underline{$0.877\pm0.071$} & \underline{$0.679\pm0.151$} & $0.934\pm0.055$ & $0.738\pm0.175$ & $0.633\pm0.065$ & \underline{$0.590\pm0.092$} \\

7 & $0.798\pm0.006$ & $0.869\pm0.036$ & $0.767\pm0.027$ & $\mathbf{0.879\pm0.045}$ & $0.673\pm0.074$ & $0.929\pm0.046$ & \underline{$0.764\pm0.100$} & $0.597\pm0.069$ & $0.573\pm0.073$ \\

9 & $0.721\pm0.008$ & \underline{$0.878\pm0.046$} & $0.764\pm0.014$ & $0.863\pm0.061$ & $0.644\pm0.105$ & $\mathbf{0.935\pm0.049}$ & $0.705\pm0.093$ & $0.571\pm0.069$ & $0.543\pm0.059$ \\

11 & $0.643\pm0.010$ & $0.856\pm0.034$ & $0.751\pm0.010$ & $0.841\pm0.075$ & $0.574\pm0.080$ & $0.931\pm0.038$ & $0.588\pm0.101$ & $0.552\pm0.049$ & $0.507\pm0.071$ \\

\bottomrule 
\end{tabular}} 
\label{tab:supervision_stress_results} 
\end{table*}

\paragraph{Clinical utility, concept quality, and spatial faithfulness}
Table~\ref{tab:main_results} compares a standard CNN classifier (EfficientNet-B2), a vanilla CBM, and the SG-CBM in terms of diagnostic performance, concept prediction, and spatial grounding. Introducing a concept bottleneck slightly reduces diagnostic performance compared to the end-to-end CNN (AUROC $0.869$ vs.\ $0.874$). However, adding spatial grounding improves performance beyond both baselines: SG-CBM achieves the highest AUROC ($0.892$) and accuracy ($0.802$), while also improving concept prediction quality (concept AUROC $0.771$ vs.\ $0.741$ for vanilla CBM). This suggests that the zone-aware weak supervision not only improves explanation quality but can also regularize concept representations.

Grounding metrics show a large improvement in localization. Without spatial supervision, CBM exhibits weak alignment with the lesion region (ROI Energy $0.149$). SG-CBM substantially increases activation concentration within target zones, achieving ROI Energy $0.815$ and Hit@1 $0.915$, while also improving posterior localization (Energy $0.585$, Top-5\% $0.567$). Figure~\ref{fig:localization_comparision} provides qualitative comparisons of concept activations. Vanilla CBM produces dispersed activations that often fall outside clinically relevant regions or are not strongly activated inside the zones. In contrast, SG-CBM concentrates responses within the lesion ROI and posterior band, consistent with the quantitative grounding metrics. Overall, SG-CBM improves diagnostic performance, concept prediction, and spatial localization simultaneously, demonstrating that weak spatial supervision can guide concept evidence toward anatomically plausible regions without requiring pixel-level concept annotations.

\paragraph{Supervision quality assessment}
A key premise of SG-CBM is that lesion delineation provides coarse but actionable supervision for spatially grounding concept evidence. In clinical practice, however, delineations may vary in quality due to annotator variability or automated segmentation errors, making supervision quality a relevant data quality factor for trustworthy AI systems. To quantify how supervision quality impacts downstream behavior in our context, we perform a Train-corrupt/Test-clean annotation-quality stress test (Table~\ref{tab:supervision_stress_results}) in which the training masks are progressively degraded using \emph{binary morphological erosion}: for each severity level, we shrink the lesion region by removing boundary pixels via a $k\times k$ sliding-window operation (single iteration), yielding systematically smaller masks (quantified by Dice {score}). During evaluation, grounding is computed using zones derived from the clean masks (Test-clean) to avoid self-consistency bias in zone-based grounding metrics and to isolate the effect of training-time supervision quality. Interestingly, Table~\ref{tab:supervision_stress_results} shows a non-monotonic sensitivity to mask quality: mild-to-moderate erosion (Dice $\approx 0.94$--0.80) generally preserves concept AUROC and diagnostic AUROC while improving several grounding indicators (e.g., ROI Energy increases from $0.815$ to $0.876$--$0.879$, and posterior grounding improves for $K=3-5$). This suggests that slightly tighter lesion cues can act as a strong regularizer, reducing activation spill and encouraging concept evidence to concentrate in clinically plausible regions. In contrast, under severe erosion (Dice $\approx\le 0.70$, $K=9-11$), both diagnostic AUROC and grounding decline, indicating a supervision-quality threshold beyond which spatial constraints can become mis-specified and reduce spatial faithfulness.

\section{Discussion \& Future Directions}
SG-CBM shows that adding anatomically motivated weak spatial supervision to a CBM can improve both explanation quality and predictive behavior. Compared with a vanilla CBM, SG-CBM yields large gains in spatial faithfulness across grounding metrics and improves diagnostic AUROC ($0.869\rightarrow0.892$), suggesting that evidence-alignment constraints can regularize concept learning and discourage reliance on background artifacts.

From a data-quality perspective, SG-CBM operationalizes task-driven data quality assurance by enforcing an evidence quality requirement: when pixel-level concept annotations are unavailable, coarse lesion delineations provide a practical supervision signal to improve the auditability of learned concept evidence. Our Train-corrupt/Test-clean annotation-quality stress test further serves as a data quality assessment methodology, demonstrating how supervision quality influences both spatial faithfulness and diagnostic behavior, and highlighting the need to characterize annotation quality when developing trustworthy healthcare AI systems.

A key design choice is the use of a posterior band, rather than a posterior region extending to the bottom of the image. In practice, a full posterior region introduces noisy supervision, as deeper tissue areas often lack meaningful acoustic patterns. Restricting supervision to a size-adaptive band immediately beneath the lesion provides more stable guidance and directs the model toward clinically plausible posterior effects. In this sense, the posterior band functions as a simple denoising mechanism, constraining the grounding signal to regions most likely to exhibit posterior acoustic phenomena while reducing the influence of irrelevant background. This is important for both trustworthiness and robustness under imperfect annotations. However, we rely on heuristic design and do not explicitly incorporate ultrasound physics in defining the posterior band. Future work should investigate physics-aware posterior modeling and uncertainty-aware zone construction, particularly for small lesions and for cases where posterior phenomena are absent or ambiguous.

From the clinical perspective, SG-CBM provides explanations that more closely match radiologic reasoning by linking diagnostic decisions to recognizable BI-RADS attributes and localizing their evidence within anatomically plausible regions. However, we assume access to radiologists' annotated lesion masks, which may introduce an additional dependency. Nonetheless, our grounding mechanisms are robust to mild-to-moderate annotation variability, and foundation models such as Segment Anything (SAM) \cite{carion2025sam3segmentconcepts} and subsequent medical imaging adaptations show promise for automated segmentation \cite{review2025Mudassar}, although integrating reliable automatic segmentation into clinical workflows remains an open challenge. A promising future direction is to integrate radiologist-in-the-loop data quality control under limited annotation budgets. For example, Active Learning (AL)-based approaches \cite{WANG2024comprehensive} could help prioritize review cases where the model exhibits low spatial faithfulness, high predictive uncertainty, prompting efficient expert correction of lesion cues (e.g., rapid mask refinement) and/or targeted concept verification. This would operationalize task-driven data quality assurance in a clinically realistic way, where comprehensive expert labeling is costly. Our findings in the supervision quality test (Table \ref{tab:supervision_stress_results}) motivate treating automatic segmentation as a data quality component of the pipeline, as downstream trustworthiness depends not only on the diagnostic model but also on the reliability of the delineations used to derive zone supervision. Incorporating segmentation quality estimation (and falling back to human review when uncertainty is high) should therefore be an important systems-level direction. 

{While SG-CBM shows that predefined anatomical zones can effectively guide concept localization, this design inherently encodes spatial priors that may bias evaluation. Specifically, activation within a zone indicates spatial alignment of evidence and should be interpreted alongside, rather than as a substitute for, semantic concept correctness \cite{knab2026whats}. Disentangling spatial plausibility from clinical validity requires dedicated studies that pair zone-based grounding with probes of semantic faithfulness. Furthermore, the mechanistic interaction between the concept bottleneck structure and zone supervision (i.e., how grounding reshapes individual concept representations and their downstream contributions) remains an open question requiring representational analysis. Finally, clinical trustworthiness ultimately requires human validation; a user study with radiologists would directly assess whether spatial grounding accelerates and strengthens expert review beyond proxy metrics.}

Our findings support a structured trustworthiness validation protocol that jointly reports clinical utility, semantic interpretability, and spatial faithfulness, thereby mitigating over-reliance on accuracy alone and promoting transparent evaluation in safety-critical clinical settings. Future work should extend this framework to larger, multi-center cohorts to assess robustness and generalization under acquisition and population shifts, as well as to other imaging domains where coarse anatomical cues provide useful spatial priors. In addition, subgroup-level analyses remain necessary to determine whether concept prediction and grounding quality vary across patient demographics, devices, or clinical sites, and whether systematic differences in annotation fidelity may affect fairness and real-world trustworthiness.

\section{Conclusion}\label{sec:conclusion}
We present a spatially grounded concept bottleneck approach for breast ultrasound diagnosis that enhances the spatial grounding of concepts, a core dimension of model trustworthiness, through task-driven evidence quality assurance. Experiments show that spatial grounding improves both spatial faithfulness and predictive performance, producing more auditable, clinically aligned explanations. To quantify the impact of supervision quality, we further conducted a stress test and found that grounding and diagnostic behavior are robust to realistic mask variations, but degrade under severe corruption, revealing a quality threshold beyond which weak supervision becomes mis-specified. Overall, these results highlight that data quality and supervision design are central determinants of healthcare AI performance and trustworthiness, and motivate the development of more robust clinical AI systems.

\bibliographystyle{IEEEtran}
\bibliography{ref}

\end{document}